\newif\ifprivate\privatetrue    
\newif\iffinal\finalfalse       

\documentclass[12pt]{amsart}
\usepackage{url}
\usepackage[mathletters]{ucs} 
\usepackage[utf8x]{inputenc} 
\usepackage[T1]{fontenc}    
\usepackage[dvipsnames]{xcolor}
\usepackage[all]{xypic}
\usepackage[margin=1.5in]{geometry}

\def\citep{\cite}
\def\[{\begin{equation}}
\def\]{\end{equation}}

\def\citet{\cite}

\def\fr#1#2{{\textstyle{#1\over#2}}}
\def\paragraph#1{{\it #1}}

\ifprivate
\def\mh#1{{\color[rgb]{.6,.3,.0}\it[[MH:\ #1]]}} 
\def\bh#1{{\color[rgb]{1.,.3,.3}\it[[BH:\ #1]]}} 
\else\def\mh#1{} 
\else\def\bh#1{} 
\fi
\newcommand{\RR}{\mathbb{R}}
\newcommand{\NN}{\mathbb{N}}

\newtheorem{definition}{Definition}[section]
\newtheorem{proposition}{Proposition}[section]

\newtheorem{theorem}{Theorem}[section]
\newcommand{\Hc}{\mathcal{H}}

\begin{document}
\title[Kernel Regression via AIT]{Bridging Algorithmic Information Theory and Machine Learning: \\ A New Approach to Kernel Learning}

\author{Boumediene Hamzi$^{1,2}$}
\address{$^1$ Department of Computing and Mathematical Sciences, Caltech, CA, USA.}
\address{$^2$ The Alan Turing Institute, London, UK.}
\email{boumediene.hamzi@gmail.com}
\author{Marcus Hutter$^{4,5}$}
\address{$^4$ Google DeepMind, London, UK. }
\address{$^5$  Australian National University, Canberra, Australia }
\email{ http://www.hutter1.net/ }
\author{Houman Owhadi$^1$}
\address{$^1$ Department of Computing and Mathematical Sciences, Caltech, CA, USA.}
\email{ owhadi@caltech.edu }

\begin{abstract}
Machine Learning (ML) and Algorithmic Information Theory (AIT) look at Complexity from different points of view. We explore the interface between AIT and Kernel Methods (that are prevalent in ML) by adopting an AIT perspective on the problem of learning kernels from data, in kernel ridge regression, through the method of Sparse Kernel Flows.  In particular, by looking at the differences and commonalities between Minimal Description Length (MDL) and Regularization in Machine Learning (RML), we prove that the method of Sparse Kernel Flows is the natural approach to adopt to learn kernels from data. This approach aligns naturally with the MDL principle, offering a more robust theoretical basis than the existing reliance on cross-validation. The study reveals that deriving Sparse Kernel Flows does not require a statistical approach; instead, one can directly engage with code-lengths and complexities, concepts central to AIT. Thereby, this approach opens the door to reformulating algorithms in machine learning using tools from AIT, with the aim of providing them a more solid theoretical foundation.

\end{abstract}


\maketitle
 %
\section{Introduction}

Algorithmic Information Theory (AIT) and Machine Learning are key approaches for analyzing complex systems. AIT aims to formalize simplicity and complexity using measures like Kolmogorov Complexity and Algorithmic Solomonoff Probability \cite{Hutter:11uiphil, Hutter:07ait, grunwald2008algorithmic}. 

Conversely, Machine Learning (ML) focuses on algorithms designed to improve performance with more data, enabling the analysis of high-dimensional complex systems, even when the model is unknown. 
Reproducing Kernels, are often used as measures of similarity in ML, and Kernel-based methods hold potential for considerable advantages  in terms of  theoretical analysis, numerical implementation, regularization, guaranteed convergence, automatization, and interpretability. Indeed, reproducing kernel Hilbert spaces (RKHS) \cite{CuckerandSmale} have provided strong mathematical foundations for analyzing dynamical systems \cite{5706920,yk1, bhcm11,bhcm1,lyap_bh,bh2020a,hamzi2019kernel, bh2020b,klus2020data,ALEXANDER2020132520,bhks,bh12,bh17,hb17,mmd_kernels_bh, akian2022learning, HAMZI2023128583} and surrogate modeling (cf. \cite{santinhaasdonk19} for a survey). Recently,  experiments by Hamzi and Owhadi and collaborators \cite{BHPhysicaD, hamzimaulikowhadi,bhkfjpl, lee2021learning, bhkfsdes, bhkfnp, bh_kfs_p4, bh_kfs_p5, bh_kfs_p6} have shown
that   Kernel Flows (KFs) \cite{Owhadi19} (an RKHS technique) can
successfully reconstruct the dynamics of  prototypical chaotic dynamical systems under both regular and irregular  sampling in time and that it can
be successful in predicting complex, large-scale systems, including climate data.

In this paper, we look at the problem of learning kernels from data from an AIT point of view and show that the problem of learning kernels from data can be viewed as a problem of compression of data. In particular, using the Minimal Description Length (MDL) principle, we show that Sparse Kernel Flows \cite{bh_kfs_p5} {  is a natural approach for learning kernels from data from an AIT point of view and that it is not necessary to use a cross-validation argument to justify its efficiency, thus giving it a more solid theoretical foundation}.

Our work can also be viewed as a  step to bridge the gap between AIT and Kernel Methods; and is to be contrasted vis-\`a-vis \cite{bach2022information}, where the author considered (Classical) Information with Kernel Methods, and our work could also be viewed as a perspective on Kernel Methods from an AIT point of view instead of the classical Information Theory point of view adopted in \cite{bach2022information}, cf. \cite{grunwald2008algorithmic} for common points and differences between AIT and the classical Information Theory. Intuitively,  the link between AIT and kernel methods is clear in the sense that since reproducing kernels are measures of similarity, then similar points can be highly compressed; and if points are dissimilar, compression is lesser\footnote{ Check Appendix \ref{app:ait_kernels} for further discussion about the link between AIT and reproducing kernels}. { This endeavor is particularly promising given the solid mathematical underpinnings of AIT, in contrast to the more restrictive theoretical foundations of classical Machine Learning.}

In Section~\ref{sec:kfs_gps} we show that the relative error used to learn the kernel in the original version of Kernel Flows can be viewed as a log-likelihood ratio. In Section~\ref{sec:MDL}, we give a brief introduction to AIT and introduce Kolmogorov Complexity (KC) and the Minimal Description/Message Length (MDL/MML) principle. In Section~\ref{sec:MDL_KFs} we establish the link between MDL and KFs,
and Section~\ref{sec:conc} concludes.
In Appendix~\ref{app:RKHS} we give a brief overview of reproducing kernel Hilbert spaces.
\section{Kernel Flows and Gaussian Processes} \label{sec:kfs_gps}

Let us consider the following Problem {\bf P}: \\
Given input/output data $(x_1, y_1),\cdots , (x_N , y_N ) \in \, \mathcal{X} \times \mathbb{R}$,  recover an unknown function $u^{\ast}$ mapping $\mathcal{X}$ to $\mathbb{R}$ such that
$u^{\ast}(x_i)=y_i$ for $i \in \, \{1,...,N\}$.

In the setting of optimal recovery \cite{owhadi_scovel_2019},  Problem {\bf P} can be turned into a well-posed problem by restricting candidates for $u$ to belong to a Banach space of functions $\mathcal{B}$ endowed with a norm $||\cdot||$ and identifying the optimal recovery as the minimizer of the relative error
\begin{equation} \label{game}
    \mbox{min}_v\mbox{max}_u \frac{||u-v||^2}{||u||^2},
\end{equation}
where the max is taken over $u \in \, \mathcal{B}$ and the min is taken over candidates in $v \in \, \mathcal{B}$ such that $v(x_i)=u(x_i)=y_i$. For the validity of the constraints $u(x_i) = y_i$,  $\mathcal{B}^{\ast}$, the dual space of $\mathcal{B}$, must contain Dirac's delta functions $\phi_i(\cdot)=\delta(\cdot-x_i)$. This problem can be stated as a game between Players I and II and can then be represented as\footnote{$v\in \, L(\Phi,\mathcal{B})$ means $v \in \, \mathcal{B}$ such that $v(x_i)=u(x_i)=y_i$.}
  \begin{equation}\label{eqdkjdhkjhffORgameban}
\text{\xymatrixcolsep{0pc}\xymatrix{
\text{(Player I)} & u\ar[dr]_{\max}\in \, \mathcal{B}    &      &v\ar[ld]^{\min}\in \, L(\Phi,\mathcal{B}) &\text{(Player II)}\\
&&\frac{\|u-v(u)\|}{\|u\|}\,.& &
}}\,
\end{equation}

If $||\cdot||$ is quadratic, i.e. $||u||^2=[Q^{-1}u,u] $ where $[\phi, u]$ stands for the duality product between $\phi \in \, \mathcal{B}^{\ast}$ and $u \in \, \mathcal{B}$ and $Q : \mathcal{B}^{\ast}\rightarrow \mathcal{B}$ is a positive symmetric linear bijection (i.e. such that $[\phi, Q \phi] \ge  0$ and $[\psi, Q \phi ] = [\phi, Q \psi]$ for $\phi,\psi \in \, \mathcal{B}^{\ast} $). In that case the optimal solution of (\ref{game}) has the explicit form \cite{owhadi_scovel_2019}
\begin{equation}\label{sol_rep}v^{\ast}=\sum_{i,j=1}^{N}u(x_i) A_{i,j} Q \phi_j, \end{equation}
where   $A=\Theta^{-1}$ and $\Theta \in \, \RR^{N \times N}$ is a Gram matrix with entries $\Theta_{i,j}=[\phi_i,Q\phi_j]$.

To recover the classical representer theorem, one defines the reproducing kernel $K$ as $$K(x,y)=[\delta(\cdot-x),Q\delta(\cdot-y)]$$
In this case, $(\mathcal{B},||\cdot ||)$ can be seen as an RKHS endowed with the norm
$$||u||^2=\mbox{sup}_{\phi \in \, \mathcal{B}^\ast}\frac{(\int \phi(x) u(x) dx)^2}{(\int  \phi(x) K(x,y) \phi(y) dx dy)}$$
and (\ref{sol_rep}) corresponds to the classical representer theorem
\begin{equation}\label{eqkjelkjefffhb}
    v^{\ast}(\cdot) = y^\top\!AK(x,\cdot),
\end{equation}
 using the vectorial notation $y^\top AK(x,\cdot)=\sum_{i,j=1}^{N}y_iA_{i,j}K(x_j,\cdot)$ with $y_i=u(x_i)$, $A=\Theta^{-1}$ and $\Theta_{i,j} =K(x_i,x_j)$.

 Now, let us consider the problem of learning the kernel from data. As introduced in \cite{Owhadi19}, the method of KFs is based on the premise that \emph{a kernel is good if there is no significant loss in accuracy in the prediction error if the number of data points is halved}. This led to the introduction of\footnote{A variant of KFs based on Lyapunov exponents in order to capture long-term behavior of the system is at \cite{BHPhysicaD}, another one based on the Maximum Mean Discrepancy that allows capturing the statistical properties of the system is at \cite{BHPhysicaD}, and another one based on the Hausdorff distance that allows reconstructing attractors is at \cite{bh_kfs_p6}.}
 \[\rho=\frac{||v^{\ast}-v^{s} ||^2}{||v^{\ast} ||^2} \]
  which is the relative error between
  $v^\ast$, the optimal recovery \eqref{eqkjelkjefffhb} of $u^\ast$ based on the full dataset
  $X^f=\{(x_1,y_1),\ldots,(x_N,y_N)\}$, and
  $v^s$  the optimal recovery  of both $u^\ast$ and $v^\ast$ based on half of the dataset $ X^s=\{(x_i,y_i)\mid i \in \, \mathcal{S}\}$ ($\operatorname{Card}(\mathcal{S})=N/2$) which admits the representation
  \begin{equation}
v^s=(y^s)^\top A^s K(x^s,\cdot)
  \end{equation}
 with $y^s=\{y_i\mid i \in \, \mathcal{S}\}$,
 $x^s=\{x_i\mid i \in \, \mathcal{S}\}$,
 $A^s=(\Theta^s)^{-1}$, $\Theta^s_{i,j}=K(x_i^s,x_j^s)$.
 This quantity  $\rho$ is directly related to the game in (\ref{eqdkjdhkjhffORgameban}) where one is minimizing the relative error of $v^{\ast}$ versus $v^s$.
Instead of using the entire the dataset $X$ one may use random subsets $X^{s_1}$ (of $X$) for $v^{\ast}$ and random subsets $ X^{s_2}$ (of $X^{s_1}$) for $v^s$.
Writing\footnote{Where $K(x,X^f)$ is the row whose entries are $K(x,x_j)$ for $x_j \in X^f$, $K(X^f,x)$ is the column whose entries are $K(x_j,x)$ for $x_j \in X^f$, and $K(X^f,X^f)$ is the matrix whose entries are $K_{i,j}=K(x_i,x_j)$ for $x_i,x_j \in X^f$. }   \begin{equation}\label{variance_rkhs}\sigma^2(x)=K(x,x)-K(x,X^f)K(X^f,X^f)^{-1}K(X^f,x)   \end{equation} we have the pointwise error bound
\begin{equation}\label{error_estimate} |u(x)-v^\ast(x)| \leq  \sigma(x) \|u\|_{\Hc},\end{equation}
 Local error estimates such as (\ref{error_estimate}) are
classical in Kriging \cite{Wu92localerror} (see also \cite{owhadi2015bayesian}[Thm. 5.1] for applications to PDEs). $\|u\|_{\Hc}$ is bounded from below (and, with sufficient data, can be approximated by) by $\sqrt{Y^{f,\top} K(X^f,X^f)^{-1} Y^f} $, i.e., the RKHS norm of the interpolant  of $v^\ast$.



The optimal recovery of $u^\ast$ based on the full dataset is  given by $v^\ast(x)=\sum_{i=1}^N c_i K(x,x_i)$  which can be represented as
a gaussian process (GP)  $$u^\ast(x) \sim GP(K(x,x),\sigma(x))$$  with  mean $K(x,x)$ and variance $\sigma^2$ given by (\ref{variance_rkhs}).

Take $X$ to be a Gaussian vector $$p(X=x)\propto \exp(- \|x\|_K^2/2)$$ where $\|x\|_K$ is the RKHS norm induced by the covariance matrix $K$ of $X$ so heuristically $$\|x\|_K^2\sim - 2 \log (p(X=x))$$

Now, consider the KF loss $\|u-v\|^2_K/\|u\|_K^2$ where $u$ interpolates the whole data and $v$ half of the data.  If we let $\xi$ be the GP with kernel $K$, and observing that $$\|u\|^2_K=\|u-v\|_K^2+\|v\|_K^2$$
then
$$\|u-v\|_K^2  \sim -2(\log p(\xi=u)-\log p(\xi=v))$$


Write $X_v$ the data points for $v$
$$v=\mathbb{E}[\xi|\xi(X_v)=f(X_v)],$$
where $f$ is the target function then $$\|u-v\|^2_K\sim-2 \log p(\xi(X_u)=f(X_u) |  \xi(X_v)=f(X_v))$$
and
$$\|u-v\|^2_K/\|u\|_K^2\sim \log p(\xi(X_u)=f(X_u) |  \xi(X_v)=f(X_v))/  \log p(\xi(X_u)=f(X_u))$$
Hence, the relative error in KFs is like log-likelihood ratio. This fact allows the application of some tools from AIT, as explained below, to show that the relative error, and consequently KFs, can be viewed as a measure of data compression in AIT.

\section{A Brief Introduction to Occam \& AIT \& MDL}\label{sec:MDL}
 Within computer science, \emph{Algorithmic Information Theory} (AIT) directly connects computation, computability theory, and information theory.  
 
The central quantity of AIT is \emph{Kolmogorov complexity}, $K(x)$, which measures the complexity of an individual object $x$ as the amount of information required to describe or generate $x$.


 If $x$ contains repeating patterns like $x=1010101010101010$ then it is easy to compress, and hence $K(x)$ will be small. On the other hand, a randomly generated bit string of length $n$ is highly unlikely to contain any significant patterns, and hence can only be described via specifying each bit separately without any compression, so that $K(x)\approx n$ bits.

\paragraph{Occam's razor.}
demands to select the simplest theory consistent with the data.
It is one of the few indispensable principles of science \citep{Hutter:11uiphil}.
Without it, complicated obscure theories explaining past data but without predictive power would thrive.
`Falsification' is not enough, and Popper's `Corroboration' is either just a new name for `Confirmation' or meaningless \cite{Good:75}.

\paragraph{Language and description length.}
Quantitative science requires quantitative definitions.
As description language (complexity) for theories, we can use program (length).
For instance, data $x=10101010$ can be explained by the program
$p_s=$`forever print(10)' or $p_l=$`forever print(10101010111111)'.
Both are consistent with $x$ but lead to different predictions.
Occam's razor tells us to prefer $p_s$ over $p_l$, since $p_s$ is shorter=simpler than $p_l$.

\paragraph{Kolmogorov complexity.}
This motivates the following universal quantification of complexity, and hence simplicity, of a string $x$.
The (Monotone) Kolmogorov complexity \citep{Hutter:08kolmo}
\begin{align*}
      K(x) ~:=~ \min_p\{\ell(p):U(p)=x...\}
\end{align*}
is the length of the shortest (non-halting) program $p$ (in bits) on a (Monotone) Universal Turing Machine (UTM)
(a minimalist general-purpose computer)
that computes binary string (starting with) $x\in \,\{0,1\}^*$.
The best explanation of $x$ is the shortest program $p^*$ for which $U(p^*)=x...$.

\paragraph{Noisy data and probabilistic models.}
In order to deal with noisy data and imperfect models,
we need to consider stochastic models=programs.
Let $\mu(x)$ be the true probability that some observation stream starts with $x$,
and ${\mathcal{M}}\ni \mu$ a class of such distributions.
Given $\mu$, the optimal code = shortest program of $x$ has length $\log[1/\mu(x)]$,
achievable e.g.\ via arithmetic encoding \cite{Hutter:24uaibook2}.
Since/if we do not know $\mu$, we can take any $\rho\in \,{\mathcal{M}}$ instead,
and code $x$ in $\log[1/\rho(x)]+K(\rho)$ bits.
Note that the arithmetic decoder needs to know $\rho$,
so we need to encode (a binary representation of) $\rho$ as well.

\paragraph{Minimal Description/Message Length Principles (MDL/MML).}

   The Minimum Description Length  (MDL) principle provides a criterion for the selection of models, regardless of their complexity, without the restrictive assumption that the data form a sample from a 'true' distribution. It can be viewed as a downscaled practical version of KC.
  
   The MDL principle recommends to use, among competing models, the one that allows to compress the data+model most. The
better the compression, the more regularity has been detected, hence the better the  predictions will be.

     The MDL principle can be regarded as a formalization of Ockham's razor, which says to select the simplest model consistent with the data.

   The  MDL principle states that the best model is the one which permits the shortest encoding of the data and the model itself. It is the addition of the code length for the model
    which separates this principle from the familiar maximum likelihood principle and makes it global in the sense that any two models, regardless of their complexity, can be compared.

 Let $M = \{Q_1,Q_2,...\}$ be a countable class of models = theories = hypotheses = probabilities over sequences ${\mathcal X}^{\infty}$, sorted w.r.t. to their
complexity $K(Q_i)$  containing the unknown true sampling distribution $\mu$. 

 For  finite X, let  $Q_i(x)$ be the $Q_i$-probability of data sequence $x \in {\mathcal X}^{\ell}$. It
is possible to code $x$ in $log(\rho(x)^{-1})$ bits, e.g. by using Huffman coding. Since $x$ is
sampled from $\mu$, this code is optimal.

 Since we
do not know $\mu$, we could select the $\rho \in M$ that leads to the shortest code on the
observed data $x$. In order to be able to reconstruct x from the code we need to
know which $\rho$ has been chosen, so we also need to code $\rho$, which takes $K(\rho)$ bits.
Hence $x$ can be coded in $\mbox{min}_{\rho \in M}{log \rho(x)^{-1}+K(\rho)}$ bits. 


Occam's razor tells us to choose the simplest model.
MDL \cite{Grunwald:07} / MML \cite{Wallace:05} quantify Occam's razor and tell us to choose the model with the shortest description:
\begin{align}\label{eq:MDLg}
  \rho^\text{MDL} ~:=~ \arg\min_{\rho\in \,{\mathcal{M}}}\{-\log \rho(x)+K(\rho)\}
\end{align}
\citet{Hutter:05mdl2px,Hutter:09mdltvp} provide fully general consistency proofs.
The expression is equivalent to penalized Maximum Likelihood (ML) estimation with the very specific penalty $K(\rho)$.
In practice, we have to replace $K(\rho)$ by more easily computable upper bounds.
For instance, if ${\mathcal{M}}=\bigcup_{d=0}^{\infty} {\mathcal{M}}_d$, where ${\mathcal{M}}_d$ is a class of i.i.d.\ distributions (e.g.\ polynomials of degree $d-1$ with Gaussian noise model), smoothly parameterized by ${\theta}\in \,{\RR}^d$, MDL becomes
\begin{align}\label{eq:MDLd}
  \rho^\text{MDL} ~:=~ \mathop{\arg\min}_{d\in \,\NN_0,{\theta}\in \,{\RR}^d}\{-\log \rho_{\theta}(x)+\fr{d}2\log(n)+O(d)\}
\end{align}
where $n=\ell(x)$.
Intuitively, for sample size $n$,
each parameter ${\theta}_i$ can (only) be estimated to accuracy $O(1/\sqrt{n})$,
hence (only) the first $\log(1/O(1/\sqrt{n}))=\fr12\log(n)+O(1)$ bits of the expansion of each of the $d$ parameters needs to be encoded.

The $O(d)$ term depends on the smoothness of the parametrization and
can explicitly be quantified as $\log\int\sqrt{\det[I({\theta})/2\pi e]}d^d {\theta}$,
where $I({\theta})$ is the Fisher Information Matrix of $\rho_{\theta}$.
Note that the dominant negative log-likelihood $-\log \rho_{\theta}(x)$ is linear in $n$,
the (BIC~\cite{neath2012bayesian}) dimensional complexity penalty $\fr{d}2\log(n)$ is logarithmic in $n$,
while $O(d)$ is a (small) curvature correction independent of $n$ that \emph{may} be acceptable to neglect.
See \cite{Wallace:05,Grunwald:07} for detailed explanations, derivations and consistency proofs for the i.i.d.\ case and \cite[Sec.6.1]{Hutter:03optisp} beyond i.i.d.

\paragraph{Bayesian derivation.}
The same expression can be derived from Bayesian principles.
If we assume priors $p_d({\theta})$ over the parameters,
\begin{align}\label{eq:MAP}
   \rho^\text{MAP} ~:=~ \mathop{\arg\max}_{{\theta}\in \,{\RR}^{d^*}}\{\rho_{\theta}(x)p_{d^*}({\theta})\},~~~\text{where}~~~
   d^* ~:=~ \mathop{\arg\max}_{d\in \,\NN_0}\{\int\rho_{\theta}(x)p_d({\theta})d^d{\theta}\}
\end{align}
Using Laplace approximation for the integral for large $n$, and taking the negative logarithm,
one can show that \eqref{eq:MAP} reduces to \eqref{eq:MDLd} \cite{neath2012bayesian}.
If we choose the reparametrization-invariant minimax-optimal Jeffreys/Bernardo prior for $p_d({\theta})$, then even the $O(d)$ term coincides with MDL.
Bayes and MDL only differ in $o(1)$ terms that vanish for $n\rightarrow{\infty}$ \cite{Wallace:05,Grunwald:07}.

  MDL is closely related to Bayesian prediction. Bayesians use $Bayes(z|x)$ for prediction, where $Bayes(x):=\sum_{
Q \in M }Q(x)w_Q $ is the Bayesian mixture with prior weights $w_Q > 0$ $\forall Q \in M$ and $\sum_{Q \in M} w_Q$. For MDL, $w_Q = 2^{-K(Q)}$.

\paragraph{Regularization in Machine Learning (RML).}
The classical approach in machine learning to combat overfitting to a too complex model is to add a regularization term to the loss function:
\begin{align*}
  f^* ~:=~ \arg\min_f\{\sum_{t=1}^n \text{loss}(f(x_t),y_t) +\lambda R(f)\}
\end{align*}
Some differences and commonalities between MDL and RML are as follows:
\begin{itemize}\parskip=0ex\parsep=0ex\itemsep=0ex
\item The MDL principle \eqref{eq:MDLg} is completely general.
It makes no assumption whatsoever on the underlying model class $\rho$:
No iid or stationarity or ergodicity or parametric or smoothness assumptions;
while RML is somewhat tailored towards iid data.
\item MDL is limited to log-loss, while any loss can be used in RML \cite{Hutter:07lorp}.
\item RML requires choosing a penalty $R$ and tuning its strength $\lambda$, often by cross-validation, while the MDL theory dictates the penalty term.
\item MDL estimates densities $\rho$, while PML determines mappings $f$,
though this difference is superficial:
One can trivially generalize $\rho(x)$ to $\rho(y|x)$ in MDL,
and augment $f$ with a noise model in PML.
\end{itemize}

\section{Kernel Flows and Algorithmic Information Theory} \label{sec:MDL_KFs}

Let $k_i(x,y; \beta)$ be some simple kernels. Our goal is to learn a kernel of the form
\[	K_{\beta,\theta} (x,y)= \sum_{i=1}^{m} \theta^2_i k_i(x,y;\beta)
\]

The problem of finding the best kernel reduces then to the following optimization problem
\[\label{combinatorial_optimization_problem}
	\mathcal{L}(\beta,\theta, p) =
	\arg\min\limits_{\beta,\theta, p}  \bigg(1 - \frac{y^\top_c K^{-1}_{\beta,\theta} y_c}{y^\top_b K^{-1}_{\beta,\theta} y_b} +
	\frac{p}{2} \log(N)\bigg),
\]
where $p=1\cdots,m$ is the number of non-zero parameters.
The $\frac{p}{2} \log(N)$ is the MDL regularization term.
In principle, one would have to solve $2^m$ optimization problems corresponding to the combinatorial problem where one solves (\ref{combinatorial_optimization_problem}) with $m-p$ out of $m$ of the $\theta_i$ set to zero. In practice the optimization problem (\ref{combinatorial_optimization_problem}) is approximated by
\[\label{approx_optimization_problem}
	\mathcal{L}(\beta,\theta) =
	\arg\min\limits_{\beta,\theta}  \bigg( 1 - \frac{y^\top_c K^{-1}_{\beta,\theta} y_c}{y^\top_b K^{-1}_{\beta,\theta} y_b} +
	\lambda ||\theta||_1  \bigg)
\]
which also penalizes each non-zero parameter, but by an amount proportional to its magnitude rather than $\frac12\log(N)$,
with the advantage of being convex. {  Replacing the MDL regularizer in (\ref{combinatorial_optimization_problem}) by the $L_1$ regularizer in (\ref{approx_optimization_problem}) is a form of \emph{surrogate loss} 
(cf. \cite{steinwart2008support} for a reference),
but for the regularization.

The general idea of surrogate loss is to replace the loss one cares about
with a different (surrogate) loss (usually convex) that is
computationally more convenient. The criterion (\ref{approx_optimization_problem}) is the one that was used in Sparse Kernel Flows introduced in \cite{bh_kfs_p5}.}


%
\section{Conclusion}\label{sec:conc}

In this study, we revisited the problem of learning kernels from data through the lens of the Minimum Description Length (MDL) principle. We demonstrated that the optimization task in Sparse Kernel Flows (SKFs) could be interpreted as a data compression problem, wherein learning kernels from data via SKFs emerges as a strategy for selecting the most concise representation, as guided by the MDL principle. This perspective has led us to validate that adopting Kernel Flows, especially Sparse Kernel Flows, for kernel learning is not only natural but also aligns with the principles of Occam's Razor, advocating for simplicity in explanatory models.

Furthermore, our findings reveal that the traditional reliance on cross-validation to establish the efficacy of KFs is not a prerequisite. This insight underscores a paradigm shift away from conventional statistical methodologies toward a direct engagement with code lengths and algorithmic complexities. We conjecture that the optimal number of data points, from an AIT point of view, corresponds to the concept of covering numbers, which we believe will yield superior model performance.

Ultimately, our broader objective is to extend this reformulation to encompass a wider array of machine learning algorithms, employing an Algorithmic Information Theory (AIT) framework to provide a more robust theoretical underpinning. This endeavor aims to not only deepen our understanding of algorithmic processes but also to enhance the theoretical foundations of machine learning techniques.

\subsection*{Acknowledgements}
BH and HO acknowledge support from the Jet Propulsion Laboratory, California Institute of Technology, under a contract with the National Aeronautics and Space Administration and from Beyond Limits (Learning Optimal Models) through CAST (The Caltech Center for Autonomous Systems and Technologies).

\appendix

\section{Reproducing Kernel Hilbert Spaces (RKHS)}\label{app:RKHS}

We give a brief overview of reproducing kernel Hilbert spaces as used in statistical learning
theory ~\cite{CuckerandSmale}. Early work developing
the theory of RKHS was undertaken by N. Aronszajn~\cite{aronszajn50reproducing}.

\begin{definition} Let  ${\mathcal H}$  be a Hilbert space of functions on a set ${\mathcal X}$.
Denote by $\langle f, g \rangle$ the inner product on ${\mathcal H}$   and let $\|f\|= \langle f, f \rangle^{1/2}$
be the norm in ${\mathcal H}$, for $f$ and $g \in \, {\mathcal H}$. We say that ${\mathcal H}$ is a reproducing kernel
Hilbert space (RKHS) if there exists a function $K:{\mathcal X} \times {\mathcal X} \rightarrow \RR$
such that\\
 i. $K_x:=K(x,\cdot)\in \, {\mathcal{H}}$ for all $x\in \, {\mathcal{X}}$.\\
ii. $K$ spans ${\mathcal H}$: ${\mathcal H}=\overline{\mbox{span}\{K_x~|~x \in \, {\mathcal X}\}}$.\\
 iii. $K$ has the {\em reproducing property}:
$\forall f \in \, {\mathcal H}$, $f(x)=\langle f,K_x \rangle$.\\
$K$ will be called a reproducing kernel of ${\mathcal H}$. ${\mathcal H}_K$  will denote the RKHS ${\mathcal H}$
with reproducing kernel  $K$ where it is convenient to explicitly note this dependence.
\end{definition}

The important properties of reproducing kernels are summarized in the following proposition.
\begin{proposition}\label{prop1} If $K$ is a reproducing kernel of a Hilbert space ${\mathcal H}$, then\\
i. $K(x,y)$ is unique.\\
ii.  $\forall x,y \in \, {\mathcal X}$, $K(x,y)=K(y,x)$ (symmetry).\\
iii. $\sum_{i,j=1}^q\beta_i\beta_jK(x_i,x_j) \ge 0$ for $\beta_i \in \, \RR$, $x_i \in \, {\mathcal X}$ and $q\in \,\mathbb{N}_+$
(positive definiteness).\\
iv. $\langle K(x,\cdot),K(y,\cdot) \rangle=K(x,y)$.
\end{proposition}
Common examples of reproducing kernels defined on a compact domain $\mathcal{X} \subset \mathrm{R}^n$ are the \\
(1) constant kernel: $K(x,y)= k > 0$ \\
(2) linear kernel: $K(x,y)=x\cdot y$ \\
(3) polynomial kernel: $K(x,y)=(1+x\cdot y)^d$ for $d\in\NN_+$ \\
(4) Laplace kernel: $K(x,y)=\exp\{-||x-y||_2/\sigma\}$, with $\sigma >0$ \\
(5)  Gaussian kernel: $K(x,y)=\exp\{-||x-y||^2_2/\sigma^2\}$, with $\sigma >0$ \\
(6) triangular kernel: $K(x,y)=\max \{0,1-||x-y||_2^2/\sigma^2\}$, with $\sigma >0$ \\
(7) locally periodic kernel: $K(x,y)=\sigma^2 \exp\{-2 \sin^2(\pi ||x-y||_2/p)/\ell^2\}\exp\{-||x-y||_2^2/\ell^2\}$, with $\sigma, \ell, p >0$.

\begin{theorem} \label{thm1}
Let $K:{\mathcal X} \times {\mathcal X} \rightarrow \RR$ be a symmetric and positive definite function. Then there
exists a Hilbert space of functions ${\mathcal H}$ defined on ${\mathcal X}$   admitting $K$ as a reproducing Kernel.
Conversely, let  ${\mathcal H}$ be a Hilbert space of functions $f: {\mathcal X} \rightarrow \RR$ satisfying
$\forall x \in \, {\mathcal X}, \exists \kappa_x>0,$ such that $|f(x)| \le \kappa_x \|f\|_{\mathcal H},
\quad \forall f \in \, {\mathcal H}. $
Then ${\mathcal H}$ has a reproducing kernel $K$.
\end{theorem}


\begin{theorem}\label{thm4}
 Let $K(x,y)$ be a positive definite kernel on a compact domain or a manifold $X$. Then there exists a Hilbert
space $\mathcal{F}$  and a function $\Phi: X \rightarrow \mathcal{F}$ such that
$$K(x,y)= \langle \Phi(x), \Phi(y) \rangle_{\mathcal{F}} \quad \mbox{for} \quad x,y \in \, X.$$
 $\Phi$ is called a feature map, and $\mathcal{F}$ a feature space\footnote{The dimension of the feature space can be infinite, for example in the case of the Gaussian kernel.}.
\end{theorem}

\section{On the link between reproducing kernels and AIT}\label{app:ait_kernels}

The intuitive connection between Kolmogorov Complexity (KC) and kernel methods readily suggests itself due to the nature of reproducing kernels as measures of similarity. Essentially, when two objects are similar, they can be efficiently compressed; conversely, dissimilar objects resist such compression. A practical way to gauge the similarity between two objects is by assessing the challenge involved in converting one into the other, which can be quantified by the length of the shortest program that transforms $y$ into $x$, denoted as $K(x|y)$. This conditional Kolmogorov Complexity, defined as  
$$
    K(x|y) := \min\nolimits_p\{\ell(p):U(y,p)=x\}
$$ 
represents the minimal program length needed to produce $x$ when given $y$ as an additional input. If $y$ offers no information about $x$ or is entirely unrelated, $K(x|y)$ will approximate $K(x)$, indicating no compression benefit from $y$. On the other hand, if $y$ encapsulates significant information pertinent to $x$, then $K(x|y)$ will be markedly less, reflecting the efficiency of information compression when relevant data is available. This connection between reproducing kernels and conditional Kolmogorov Complexity opens the door to the possibility of approximating the  Kolmogorov Complexity using kernel methods that we leave for future work.

\bibliographystyle{alpha}
\bibliography{bibliography, missing_dyn}

\end{document}